\documentclass[10pt,twocolumn,letterpaper]{article}
\pdfoutput=1
\usepackage{cvpr}
\usepackage{times}
\usepackage{epsfig}
\usepackage{graphicx}
\usepackage{amsmath}
\usepackage{amssymb}
\usepackage{float}

\usepackage[pagebackref=true,breaklinks=true,letterpaper=true,colorlinks,bookmarks=false]{hyperref}
\usepackage[breaklinks=true,bookmarks=false]{hyperref} 

\cvprfinalcopy 

\def\cvprPaperID{1549} 

\newcommand{\argmax}{\operatornamewithlimits{arg\,max}}

\ifcvprfinal\fi
\begin{document}

\title{When will you do what? - Anticipating Temporal Occurrences of Activities}

\author{Yazan Abu Farha, Alexander Richard, Juergen Gall\\
University of Bonn, Germany\\
{\tt\small \{abufarha,richard,gall\}@iai.uni-bonn.de}
}

\newcommand{\TODO}[1]{\textcolor{red}{\textbf{TODO: }#1}}

\maketitle

\begin{abstract}
    Analyzing human actions in videos has gained increased attention recently.
    While most works focus on classifying and labeling observed video frames
    or anticipating the very recent future, making long-term predictions
    over more than just a few seconds is a task with many practical applications
    that has not yet been addressed.
    In this paper, we propose two methods to predict a considerably large amount
    of future actions and their durations. Both, a CNN and an RNN are trained
    to learn future video labels based on previously seen content. We show
    that our methods generate accurate predictions of the future even for long
    videos with a huge amount of different actions and can even deal with
    noisy or erroneous input information.
\end{abstract}


\section{Introduction}
\label{sec:introduction}
In the last years, we have seen a tremendous progress in the capabilities of computer systems to classify and segment activities in videos, \eg~\cite{shou,rnn_model,lea}. These systems, however, analyze the past or in the case of real-time systems the present with a delay of a few milliseconds. For applications, where a moving system has to react or interact with humans, this is insufficient. For instance, collaborative robots that work closely with humans have to anticipate the activities of a human in the future. In contrast to humans that are very good in anticipating activities, developing methods that anticipate future activities from video data is very challenging and has just recently received an increase of interest. 

Current works anticipate activities only for a very short time horizon of a few seconds. While early activity detection addresses the problem of inferring the class label of an action at the point when the activity starts or shortly thereafter~\cite{early_1,early_2,early_3,early_4}, other works predict the class label of the action that will happen next~\cite{Pei:2013:LPV,Jain_2016_CVPR,Lan}. In the recent work \cite{iccv17}, the starting time of the future activity is estimated as well.

In this work, we go beyond the recognition of an ongoing activity or the anticipation of the next activity. We address the problem of anticipating all activities that will be happening within a time horizon of up to 5 minutes. This includes the classes and order of the activities that will occur as well as when each activity will start and end. Figure~\ref{fig:qual} shows a few example predictions. 

To address this problem, we propose two novel approaches for this task. In both cases, we first infer the activities from the observed part of the video using an RNN-HMM~\cite{rnn_model}. The first approach builds on a recurrent neural network (RNN) that predicts for a given sequence of inferred activities the remaining duration of the ongoing activity as well as the duration and class of the next activity. The anticipated activities are then fed back to the RNN in order to anticipate activities for a longer time horizon. The second approach builds on a convolutional neural network (CNN). To this end, we convert the temporal sequence of inferred activities in a matrix that encodes both the
length and the action label information. The CNN then predicts a matrix that encodes the length and the action labels of the anticipated activities. In contrast to the RNN approach, the CNN approach anticipates all activities in one pass.              

We have evaluated the two approaches on two challenging datasets that contain long sequences and large variations. Both approaches outperform by a large margin two baselines, a grammar based baseline and a nearest neighbor baseline. 
While the RNN and CNN perform similarly for a long time horizon of more than 40 seconds, 
the RNN performs better for shorter time horizons less than 20 seconds. 
Both approaches also outperform the method of~\cite{vondrick} that does not anticipate activities directly but visual representations of future frames, which can then be used to classify the activities.

\section{Related Work}
\label{sec:relatedWork}

Predicting future frames, poses or image segmentations in videos has been studied in several works~\cite{ranzato,multi_scale,luc_iccv,vond_gen,dual_gan,sriv,spatio,vill,walker_iccv,vuk,oh,lotter}. Approaches that predict future frames at a pixel-level, however, are limited to a few frames. Instead of predicting frames, a deep network is trained in  \cite{vondrick} to predict visual representations of future frames. The predicted visual representations can then be used to classify actions or objects using standard classifiers. This approach, however, is also limited to a very short time horizon of only 5 seconds.   
This work has been extended in \cite{red}, where they use an encoder-decoder network to predict a sequence of future representations based on a history of previous frames. 
To get the predicted actions, the output of the encoder-decoder network is passed through another network that generates the action labels. 

The task of early activity detection is also related, but it assumes that a partial observation of the ongoing activity is available, and the goal is to recognize this activity with the least possible amount of observations~\cite{early_1,early_2}. 
Recent approaches for this task use Long Short-Term Memory (LSTM) networks with special loss functions that encourage early activity detection~\cite{early_3, early_4}. 

A slightly longer time horizon is considered for approaches that predict the next action that will happen. \cite{Lan} predict future actions from hierarchical representations of short clips or still images. They encode the observed frames in a multi-granular hierarchical representation of movements, and train different classifiers at each level in the hierarchy in a max-margin framework. Recently, \cite{iccv17} train a deep network to predict the future activity and its starting time from a sequence of preceding activities. Their model relies on appearance based and motion based features extracted from the observed activities to predict what will happen next in the future. \cite{on_wrist} combine on-wrist motion sensing and visual observations to anticipate daily intentions. 
In \cite{koppula}, observed activities are modeled with spatio-temporal graphs which are used for anticipating object affordances, trajectories, and sub-activities. 
Besides of activities, anticipating goal states from first person videos has also been addressed.  
\cite{fpv} model the human behavior with a Markov Decision Process (MDP) and use inverse reinforcement learning to infer all elements of the MDP  and the reward function from first person vision videos. Then, they use the learned MDP to anticipate goal states and the length of trajectories towards them. Inverse reinforcement learning is also used in \cite{visual_forecasting} for visual sequence forecasting.


Other works investigate future prediction in the sports domain. For example, \cite{sport} use augmented hidden conditional random field to predict the location of future events in sports, like predicting shot location in tennis. Recently, a framework has been introduced in \cite{sport_iccv} to predict the next move of players or the location of the ball in the immediate future from sports videos. 

In contrast to previous work that anticipate the next activity only within a short time horizon of a couple of seconds, we address the problem of anticipating a sequence of activities including the start and end points within time horizons of up to 5 minutes.  
\section{Anticipating Activities}
\label{sec:technicalDetails}


Our goal is to anticipate from an observed video what will happen next in the video for a given time horizon, which can take up to 5 minutes. As shown in Figure~\ref{fig:qual}, we aim to predict for each frame in the future the label of the activity that will happen.     
More formally, let $ \mathbf{x}_1^T = (x_1,\dots,x_T) $ be a video with $ T $ frames.
Given the first $ t $ frames $ \mathbf{x}_1^t $, the task is to predict the actions
happening from frame $ t+1 $ to $ T $. That is, we aim to assign action labels
$ \mathbf{c}_{t+1}^T = (c_{t+1},\dots,c_T) $ to each of the unobserved frames. 


%
%

\subsection{Inferring Observed Activities}

\begin{figure}[tb]
    \centering
    \includegraphics[scale=0.18]{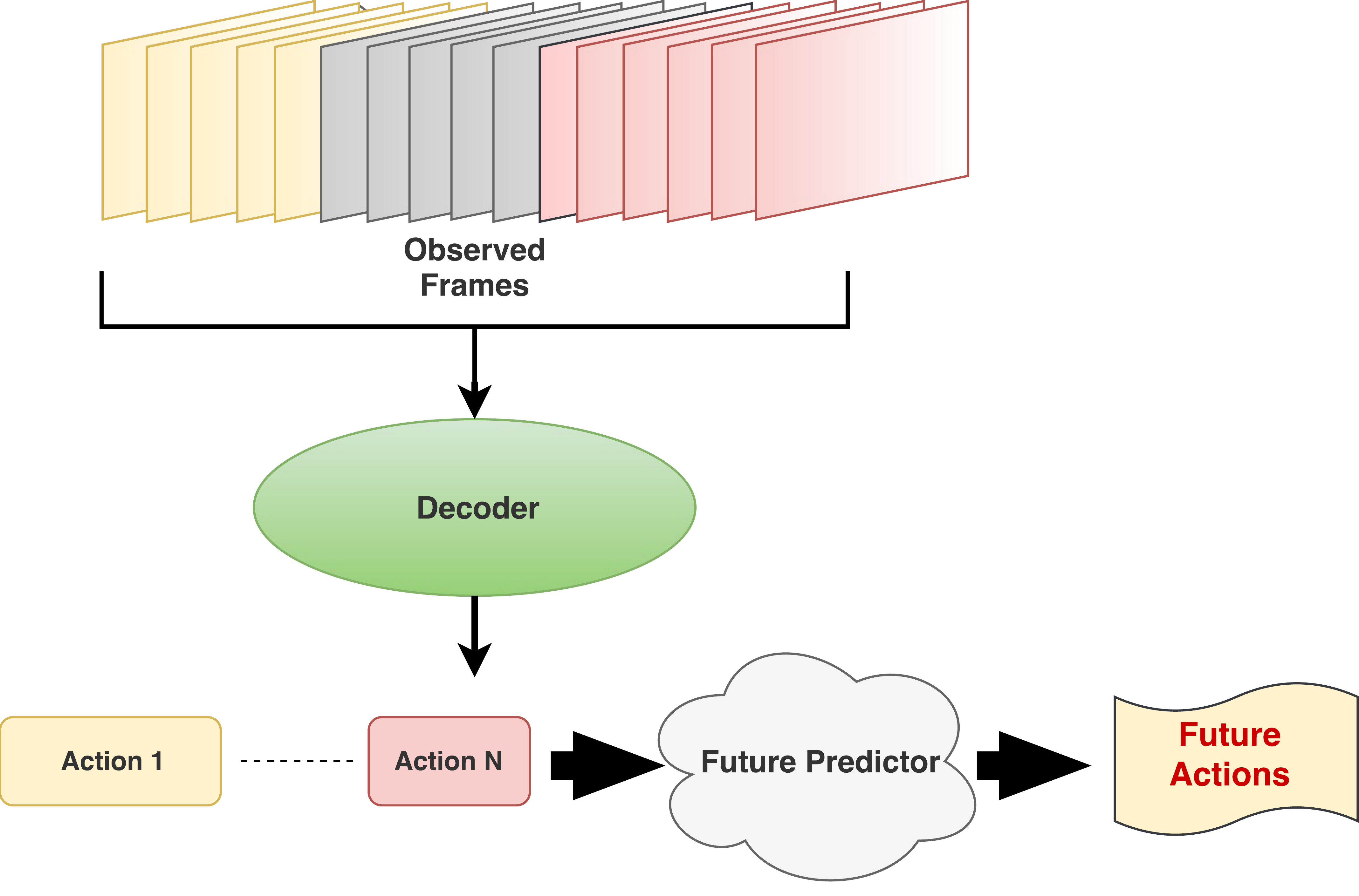}
    \caption{Proposed approach for future action prediction. From the observed frames,
             action labels are inferred by a decoder. 
             The future predictor predicts from the inferred frame labels $ \mathbf{c}_1^t $ the labels $ \mathbf{c}_{t+1}^T $ that are yet to come.}
    \label{fig:twoStep}
\end{figure}

Instead of predicting the future actions $ \mathbf{c}_{t+1}^T $ directly from the video frames
$ \mathbf{x}_1^t $, we first infer the actions $ c_1,\dots,c_t $ for the given frames $ x_1,\dots,x_t $ and then predict the future actions $ c_{t+1},\dots,c_T  $ from the inferred actions $ \mathbf{c}_{1}^t $ as it is illustrated in Figure~\ref{fig:twoStep}. This has the advantage that we can separately study the impact of the network, which infers activities from observed video sequences, and the network that anticipates the future activities. In our experiments, we will also show that the predictor network performs worse if activities are directly anticipated from the observed video frames.       

For inferring the activities $ \mathbf{c}_{1}^t $ from $ \mathbf{x}_1^t $, we use a hybrid RNN-HMM approach~\cite{rnn_model}. 
In contrast to~\cite{rnn_model}, which train the method in a weakly supervised setting, we train the model fully supervised since in our training set each video frame $ x_t $ is labeled with a class $ c_t $.

For inferring the activities $ \mathbf{c}_{t+1}^T $ from $ \mathbf{c}_1^t $, we investigate two architectures. The first architecture is based on a recurrent neural network (RNN) and will be described in Section~\ref{sec:rnn}. The second architecture is based on a convolutional neural network (CNN) and will be described in Section~\ref{sec:cnn}. 

The source code for both the RNN and CNN models is publicly available at
\url{https://github.com/yabufarha/anticipating-activities}.

\subsection{RNN-based Anticipation}
\label{sec:rnn}

\begin{figure}[tb]
    \centering
    \includegraphics[width=\columnwidth]{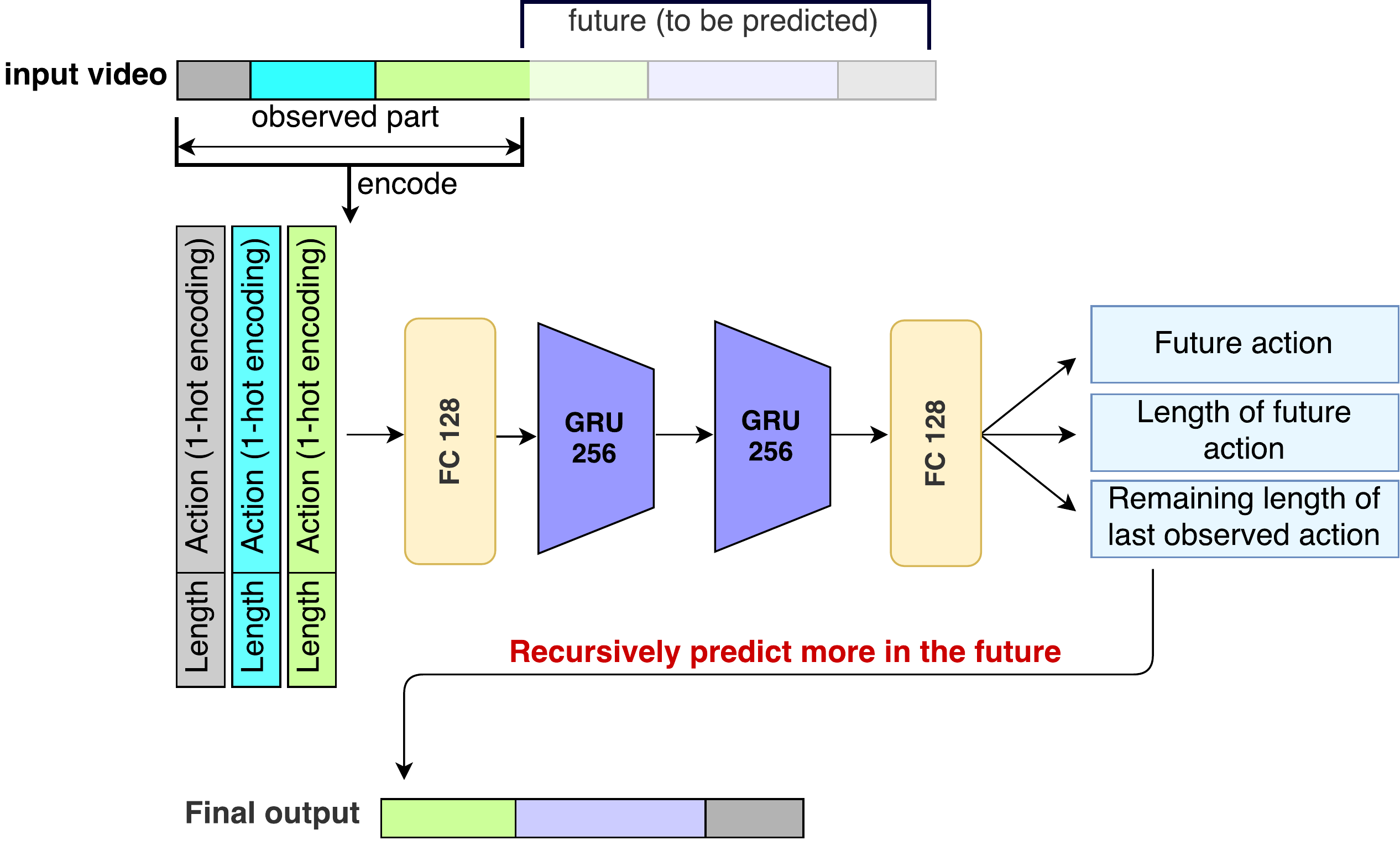}
    \caption{Architecture of the RNN system. The input is a sequence of \textit{(length, 1-hot class encoding)}-tuples.
             The network predicts the remaining length of the last observed action and the label and length of the next action.
             Appending the predicted result to the original input, the next action segment can be predicted.}
    \label{fig:rnn}
    \vspace{2mm} 
\end{figure}

\begin{figure}[tb]
    \centering
    \includegraphics[width=\columnwidth]{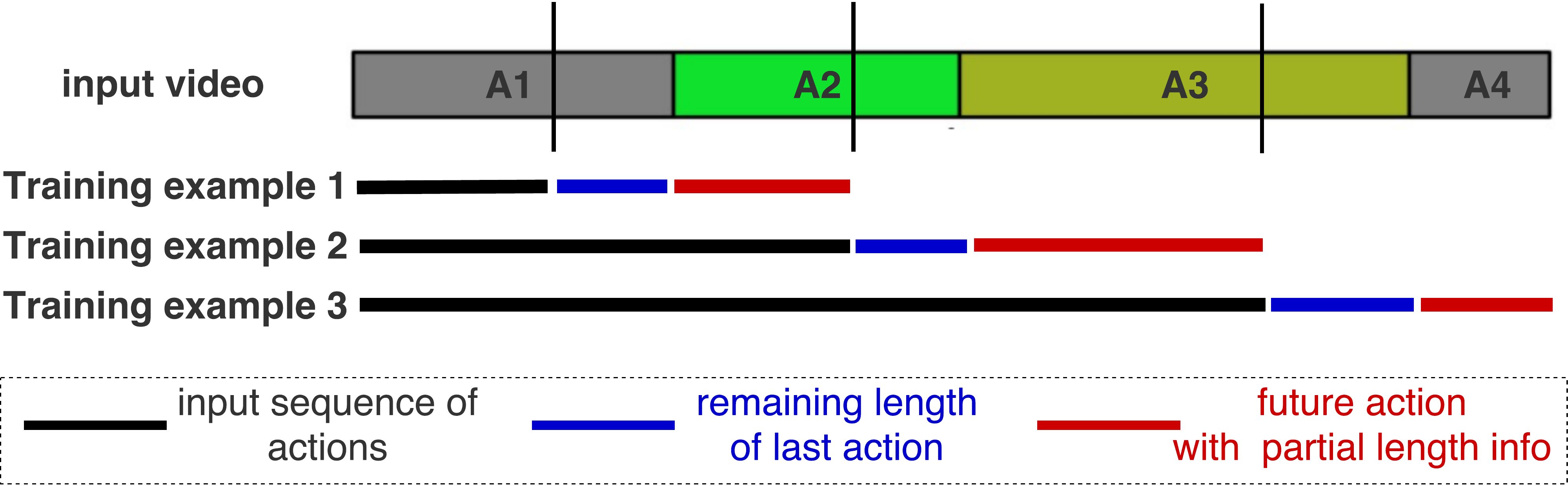}
    \caption{Training data is generated by cutting the ground-truth segmentations at random points
             and using the left part as input and the next action segment to the right of the cut
             as ground-truth for the prediction.}
    \label{fig:rnn_training}
    \vspace{2mm} 
\end{figure}

We can interpret future action prediction as a recursive sequence prediction:
As input sequence, the RNN obtains all observed segments and predicts the remainder of the
last segment as well as the next segment. This is repeated until the desired amount of
future frames is reached.

More precisely, for each observed segment, the RNN gets its class label in form of a
$ 1 $-hot encoding and the corresponding segment length, which is normalized by the video length, as input. Sequentially forwarding
all those segments, three output predictions are made: the remaining length of the last
observed segment as well as a label and a length for the next segment.
This prediction is concatenated with the observed segments to form a new input for the
network. The new input is again forwarded through the network to produce the next
prediction. The final result is obtained by repeatedly forwarding the previously
generated prediction until the desired amount of frames is predicted, see Figure~\ref{fig:rnn}.

As RNN architecture, we use two stacked layers of $ 256 $ gated recurrent units
and fully connected layers at the input and output. As output layer for both length predictions, remaining length of current action and length of next action, we use a rectified linear unit to ensure positive length outputs. 
The label prediction is done via a softmax layer as usual for classification tasks.

\paragraph{RNN Training.}
The training data generation for the RNN is illustrated in Figure~\ref{fig:rnn_training}.
Given a ground-truth labeling of a training sequence with $ n $ action segments,
$ n-1 $ training examples are generated out of it.
For a segment $ i < n $, a random split point is defined. Everything
before this point is encoded as a sequence of $ i $ tuples containing the length of
the observed segment and its label as $ 1 $-hot encoding. Each such sequence
is an input training example for the network.
For segment $ i+1 $, another random split point is defined. The values between the
first and second split point define the target the network should predict: A triplet consisting
of the remaining length of segment $i$ ($ l_r $), the length of the next
action $i+1$ from its start up to the split point ($l_n$), and the label of the next action ($c$).
By processing each training sequence like this, a large amount of input tuple sequences and
target triplets is generated. 

As loss for a single training example, we use
\begin{equation}
\mathcal{L} = - \log{\hat{p}_{c}} + (l_r-\hat{l}_r)^2 + (l_{n}-\hat{l}_n)^2, 
\end{equation}
where $\hat{l}_r$ denotes the predicted remaining length of the current action, $\hat{l}_n$ denotes the predicted length of the next
action, and $\hat{p}_{c}$ the predicted class probability of the next action. For training, we minimize the loss, which is summed over all training examples, by backpropagation through time.


\subsection{CNN-based Anticipation}
\label{sec:cnn}

\begin{figure*}[tb]
    \centering
    \includegraphics[scale=0.4]{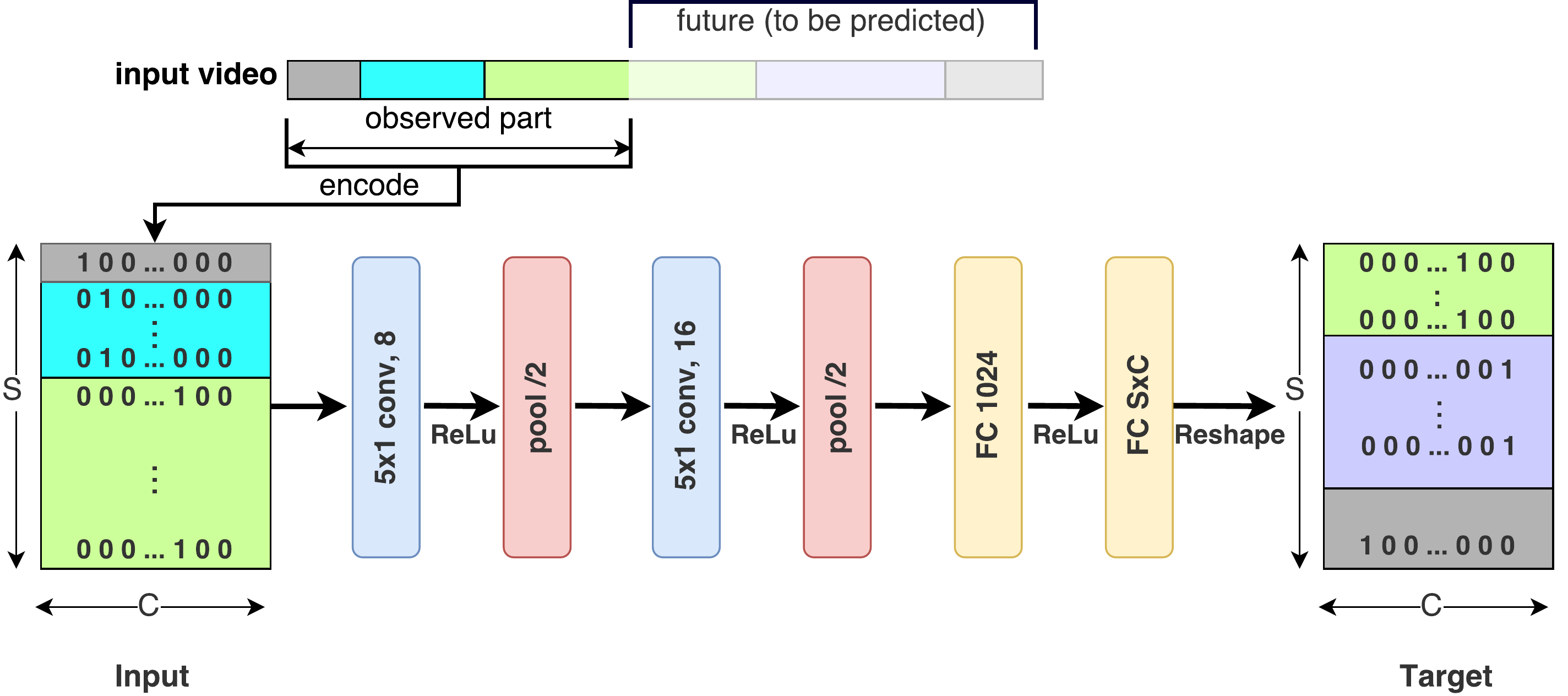}
    \caption{Architecture of the CNN-based anticipation approach. Both the input sequence and output sequence are converted into a matrix form where $C$ denotes the number of classes and $S$ corresponds to the number of video segments of a certain length. The binary values of the matrix indicate the label of each video segment.     
    }
    \label{fig:cnn}
\end{figure*}

The CNN-based anticipation approach aims at predicting all actions directly in one single step, rather
than relying on a recursive strategy such as the RNN.
The given framewise labels $ \mathbf{c}_1^t $ are encoded in a matrix $X$ with $C$ columns and $S$ rows. While the columns correspond to the $C$ action classes, the rows correspond to action segments. The number of rows for an action segment of length $l$ is given by $ \lfloor \frac{l}{t}S \rfloor$ and, for each row $s$, $X_{sc} = 1$ for the label $c$ of the corresponding action segment and zero otherwise. The matrix is filled in the order the actions occur as illustrated in Figure~\ref{fig:cnn}. We set $S$ large enough such that each action segment covers at least one row.             


The matrix $X$ that encodes the observed labels $ \mathbf{c}_1^t $ is forwarded through a CNN which consists of two convolutional layers and two fully connected layers. The convolutional layers have $ 8 $ and $ 16 $ feature maps respectively and perform a $ 5 \times 1 $ convolution followed by a rectified linear unit and max pooling. After the fully connected layers, the output layer is reshaped to a matrix $Y$, which has the same size as matrix $X$, and each row of $ Y $ is $ \ell_2 $ normalized. We further apply a 1D Gaussian filter along each column for temporal smoothing. We will show in the experiments that the additional smoothing reduces the effect of spurious predictions of very short segments, see Figure~\ref{fig:post_proc}. To convert $Y$ back into a sequence of labels $\mathbf{c}_{t+1}^T$,
we compute for each row 
\begin{align}
    \hat{c}_s = \argmax_{c}Y_{sc}
\end{align}
and concatenate $\hat{c}_s$ where each row corresponds to $\lfloor \frac{T-t}{S} \rfloor$ frames.

\paragraph{CNN Training.}

Since the CNN approach predicts all actions directly while the RNN uses a recursive strategy, we also have to prepare the training data slightly differently.   
For each video in the training set, we generate $4$ training examples by using the first $10\%$, $20\%$, $30\%$, and $50\%$ of the video, respectively, as observation and the following $50\%$ of the video as ground-truth for the prediction. For each training example, we convert the sequence of action labels $\mathbf{c}_1^t $ into the matrix $X$ and the labels of the following $50\%$ of the video frames into the ground-truth matrix $Y$. 
To train the network, we use the squared error criterion over all output elements
\begin{align}
    \mathcal{L} = \frac{1}{SC} \sum_{s,c} (Y_{sc} - \hat{Y}_{sc})^2,
\end{align}
where $\hat{Y}$ is the prediction of the network. 
We found this loss in combination with the row-wise $ \ell_2 $-normalization of the output more robust than a standard softmax output with cross-entropy loss which we attribute to the smoothing properties of the softmax function, see Section~\ref{sec:experiments} for an evaluation.

\section{Experiments}
\label{sec:experiments}

\subsection{Setup}

We conduct our experiments on two benchmark datasets for action recognition.
The \textbf{Breakfast} dataset \cite{breakfast} contains $ 1,712 $ videos of $ 52 $ different actors making breakfast. Overall, there are $ 48 $ fine-grained action classes and about $ 6 $ action instances for each video. The average duration of the videos is $ 2.3 $ minutes. We use the four splits as proposed in \cite{breakfast}.

The \textbf{50Salads} dataset \cite{50salads} contains $ 50 $ videos with $ 17 $ fine-grained action classes. With an average length of $ 6.4 $ minutes, the videos are comparably long and contain $ 20 $ action instances per video on average. Following \cite{50salads}, we use five-fold cross-validation.

The longest video in both datasets is 10 minutes. As evaluation metric, we report the accuracy of the predicted frames as mean over classes (MoC).

\paragraph{Video Representation.}
We evaluate our systems on two settings. The first is with ground-truth observations, \ie the observed labels are the ground truth annotation. This setting allows for a clean analysis of the prediction capabilities of our systems.
As a second setting, we consider the labels of the observed part of the videos to be obtained using the decoder from~\cite{rnn_model}. This way, already the observed labels can contain errors and prediction is much harder as previous errors are propagated into the future. In order to obtain the decoded labels from~\cite{rnn_model}, we compute improved dense trajectories over the observed frames $ \mathbf{x}_1^t $ and then reduce the dimensionality to 64 using PCA. After that, Fisher vectors are computed for each frame using a temporal window of size $ 20 $. For Fisher vectors computation, a Gaussian mixture model (GMM) with 64 Gaussians is built using $ 150,000 $ random samples of the dense trajectory features. Power- and $\ell_2$-normalization are applied, and at the end, the dimensionality is again reduced to 64 using PCA.

\paragraph{Parameters.}
%

For the CNN approach, we set the number of rows $S$ of the matrix $X$ to 128 for Breakfast. This ensures that each action segment covers at least one row. Since the average video length for 50Salads is about four times larger than for Breakfast, we use 512 for 50Salads.
Since increasing $S$ and therefore sampling at a finer temporal resolution did not significantly change the results, we stick to these values for the remainder of the paper. For the post-processing, we use Gaussian smoothing with $\sigma=3$ for Breakfast
and $\sigma=13$ for 50Salads. 
For the RNN approach, the normalized length input is scaled by the average number of actions in the videos 
to ensure numerical stability. 
In all experiments, the Adam optimizer is used  with a learning rate of $0.001$.

\paragraph{Baselines.}
We define two baselines for future action prediction to compare against our two proposed systems.
The first baseline uses a grammar and the mean length of each action class.
The mean length of each action class is estimated from the training data. 
The finite grammar generates all action sequences that have been observed during training. 
This is the same grammar as proposed in~\cite{rnn_model}.
Given the observed actions $ \mathbf{c}_1^t $, either based on the ground truth or on the action decoder~\cite{rnn_model},
we randomly select an action sequence from the grammar that has $ \mathbf{c}_1^t $ as prefix.
We then predict action labels $ \mathbf{c}_{t+1}^T $ such that the labels are consistent with the chosen action sequence of the grammar.
Each action class from the chosen grammar path that has not been observed, \ie that is to appear in the future,
is added with its mean class length to the prediction until all required frames from $ t+1 $ to $ T $
are predicted.

As a second baseline, we use a nearest neighbor approach. We search the nearest neighbor in the training set using 
frame-wise error of the observed part as distance, and use the remaining part as future prediction.

\subsection{Prediction with Ground-Truth Observations}

In order to provide a fair evaluation, we first assume that the observed segmentation $ \mathbf{c}_1^t $
of the video is perfect, \ie that the observed labels do not contain any errors. While this is not the
case in most realistic settings, it allows to get a clean evaluation how well the systems can predict
the future. With noisy observations, the results are more delusive as errors in the observed part
are propagated to the future.

\begin{figure}[t]
\centering
\begin{tabular}{cc}
\includegraphics[width=.45\columnwidth]{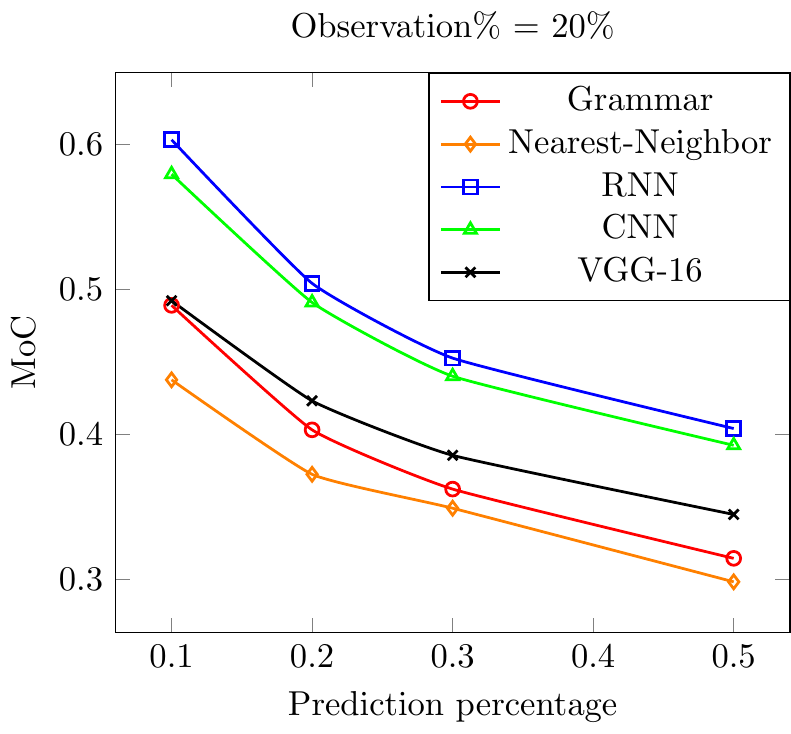}
&
\includegraphics[width=.45\columnwidth]{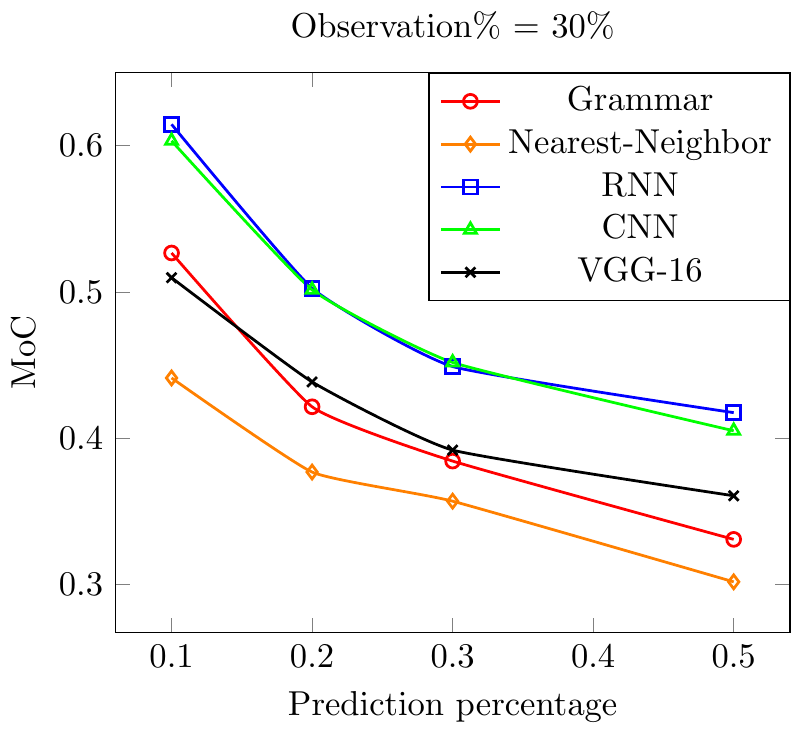} 
\\

\multicolumn{2}{c}{
\begin{footnotesize}
(a) Results on Breakfast
\end{footnotesize}
}

\\
\includegraphics[width=.45\columnwidth]{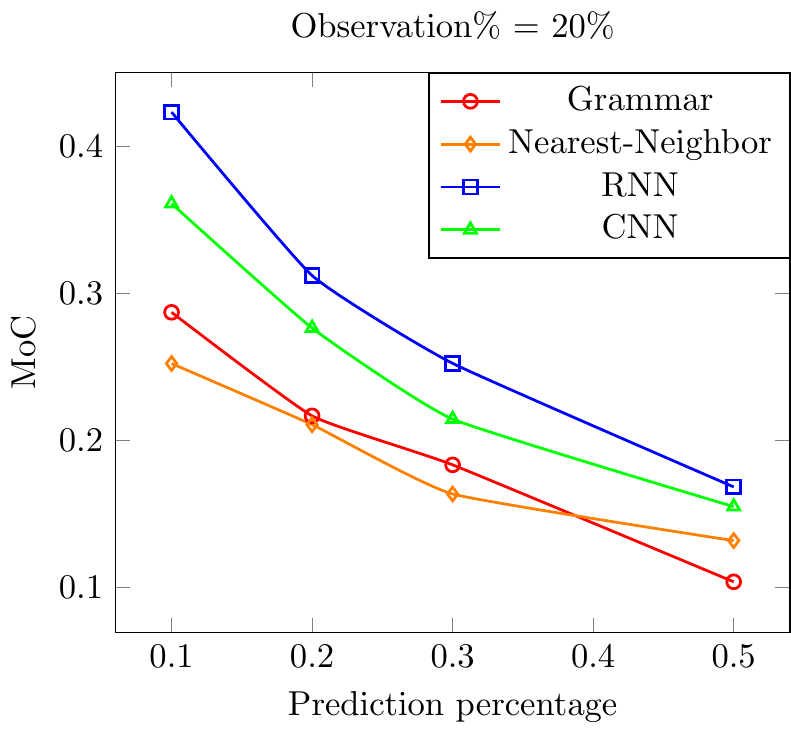}
&
\includegraphics[width=.45\columnwidth]{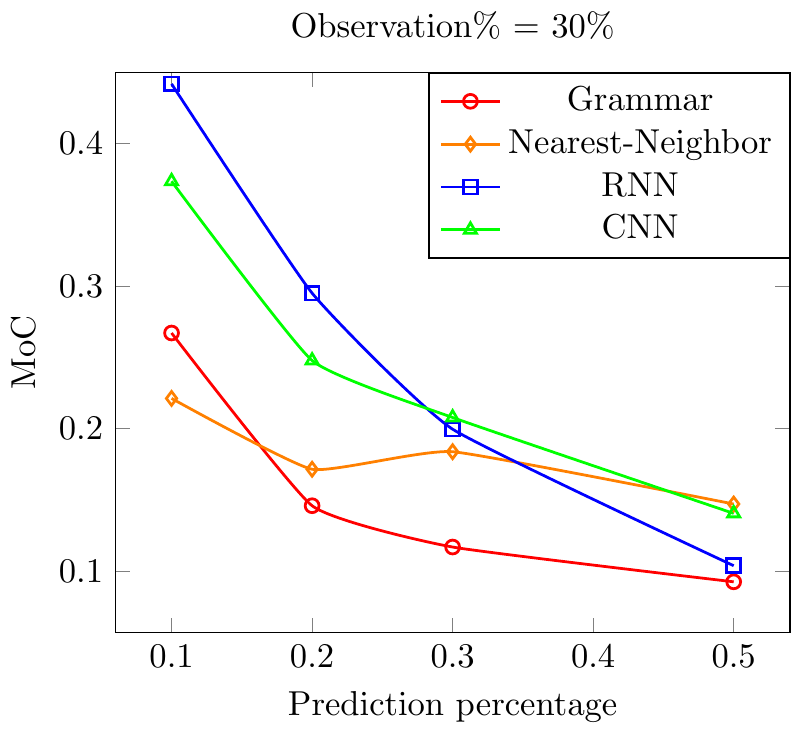} 
\\
\multicolumn{2}{c}{
\begin{footnotesize}
(b) Results on 50Salads
\end{footnotesize}
}
\end{tabular}
\caption{Results for future action prediction with ground-truth observations. }
\label{fig:plots_gt}
\vspace{-2mm} 
\end{figure}

\begin{figure*}[t]
\centering
\begin{tabular}{cc}
\includegraphics[trim={3mm 5cm 3mm 3cm},clip,width=.4\textwidth]{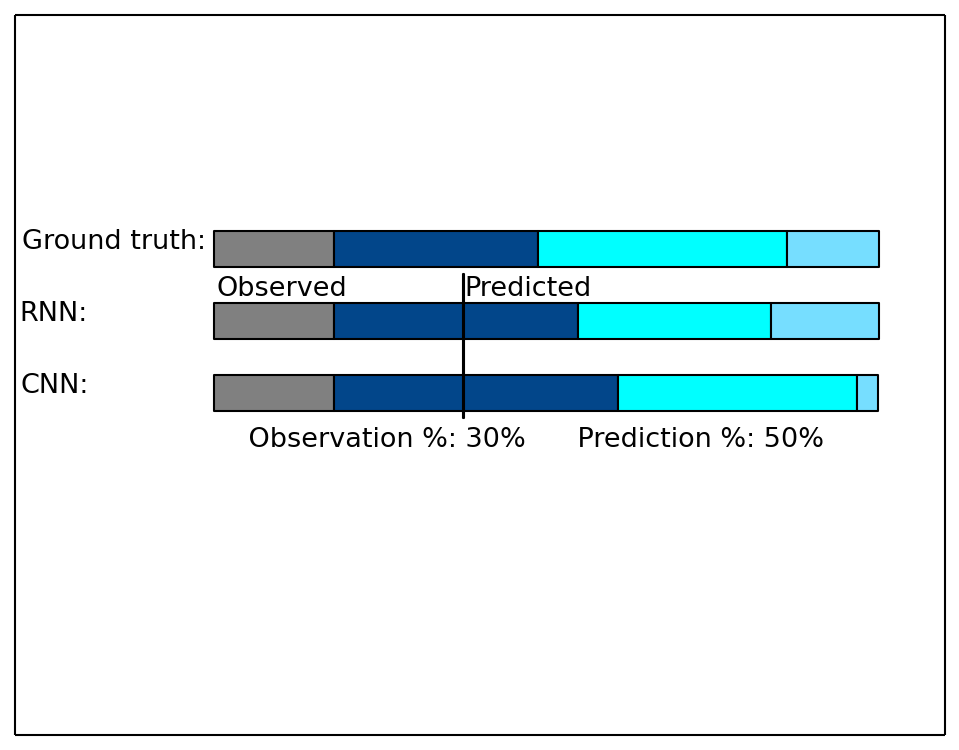} 
& 
\includegraphics[trim={3mm 5cm 3mm 3cm},clip,width=.4\textwidth]{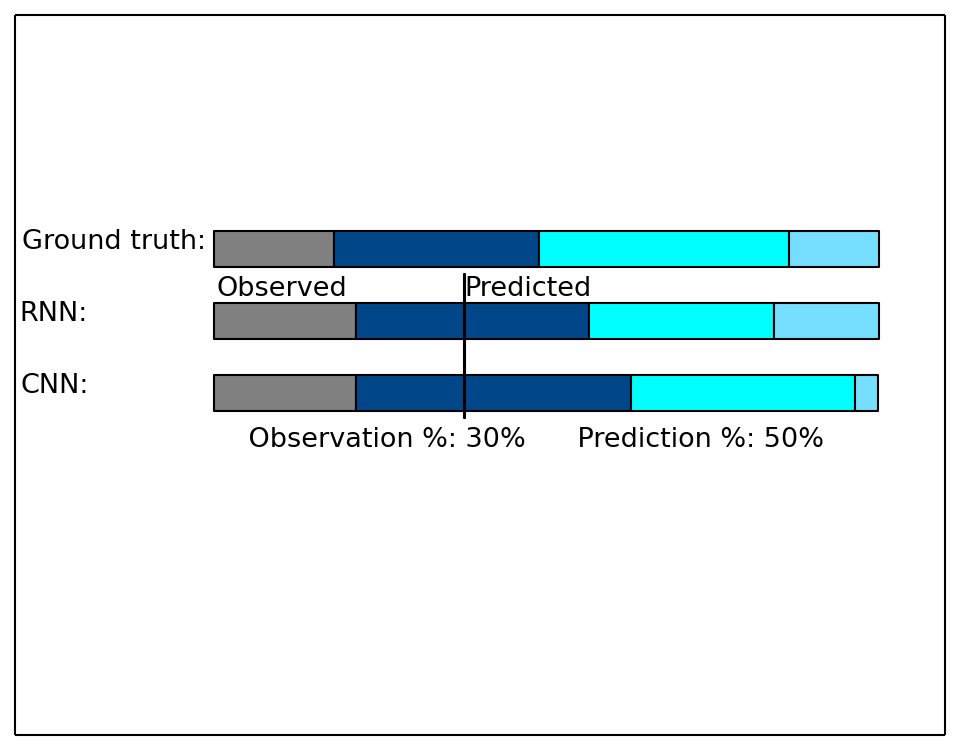}
\\
\includegraphics[trim={3mm 5cm 3mm 3cm},clip,width=.4\textwidth]{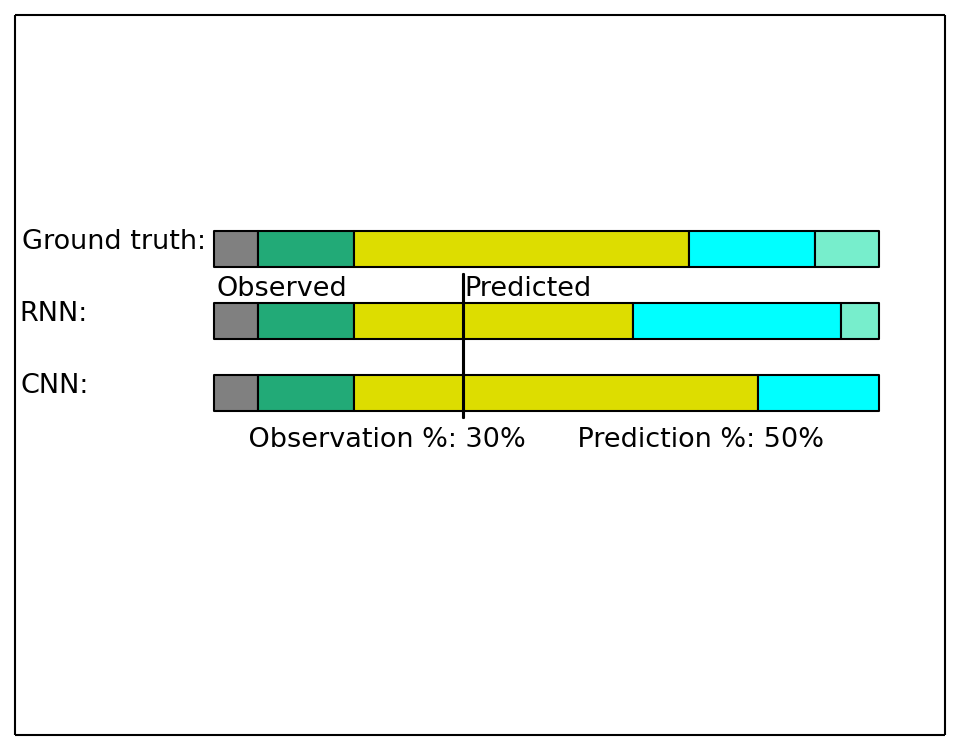} 
& 
\includegraphics[trim={3mm 5cm 3mm 3cm},clip,width=.4\textwidth]{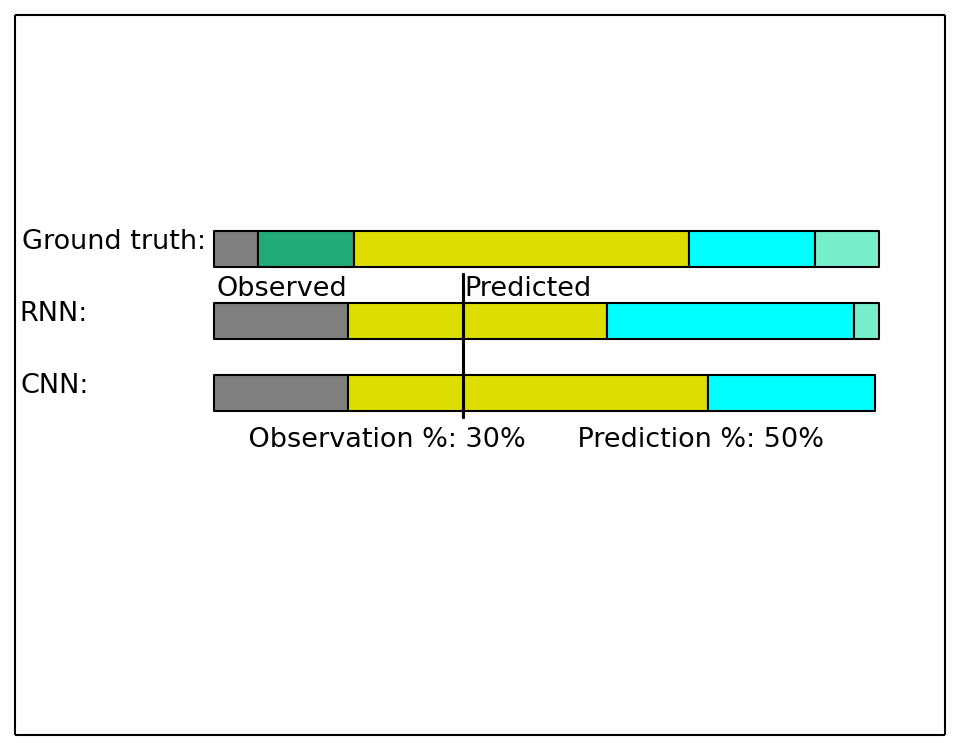}
\\
\begin{footnotesize}(a) Results on Breakfast with ground-truth observation \end{footnotesize}
& 
\begin{footnotesize}(b) Results on Breakfast without ground-truth observation \end{footnotesize}
\\
\includegraphics[trim={3mm 5cm 3mm 3cm},clip,width=.4\textwidth]{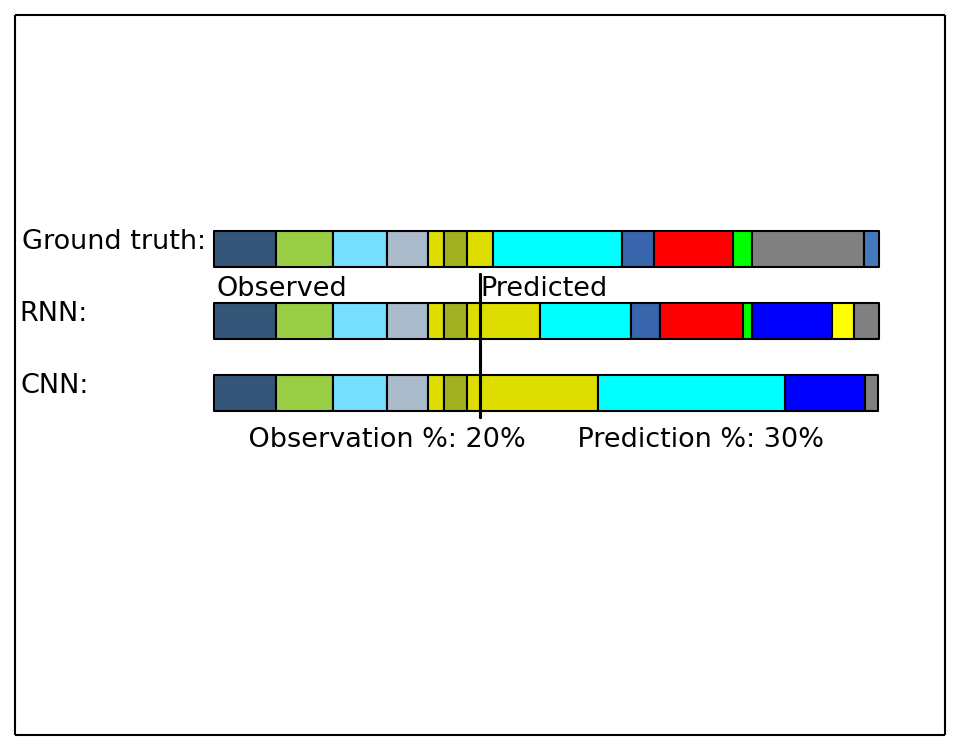} 
& 
\includegraphics[trim={3mm 5cm 3mm 3cm},clip,width=.4\textwidth]{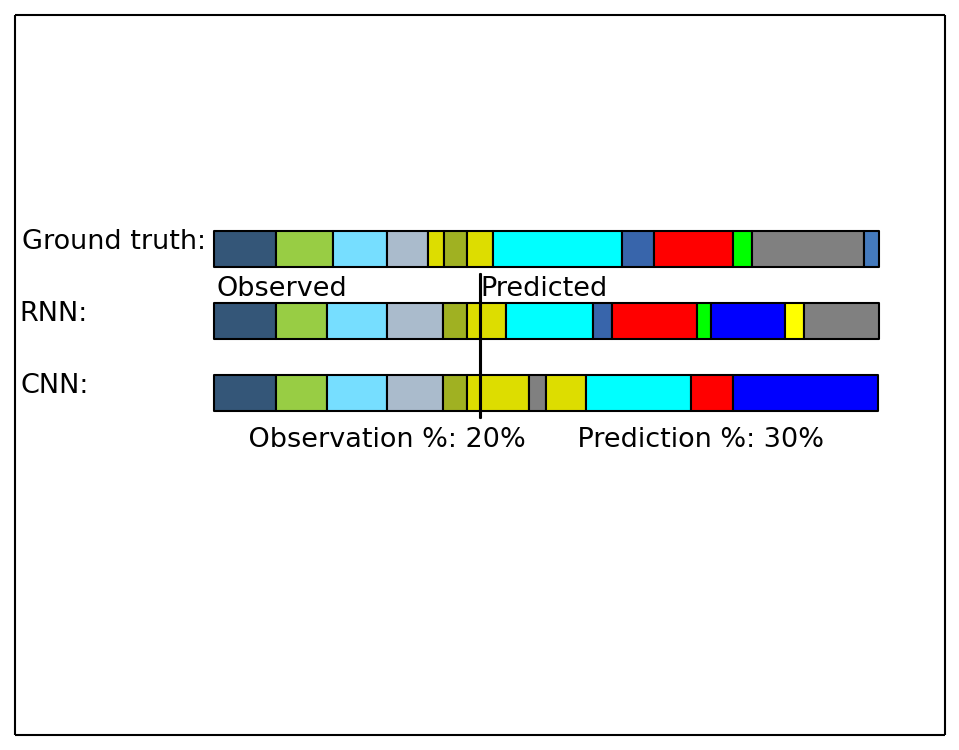}
\\
\includegraphics[trim={3mm 5cm 3mm 3cm},clip,width=.4\textwidth]{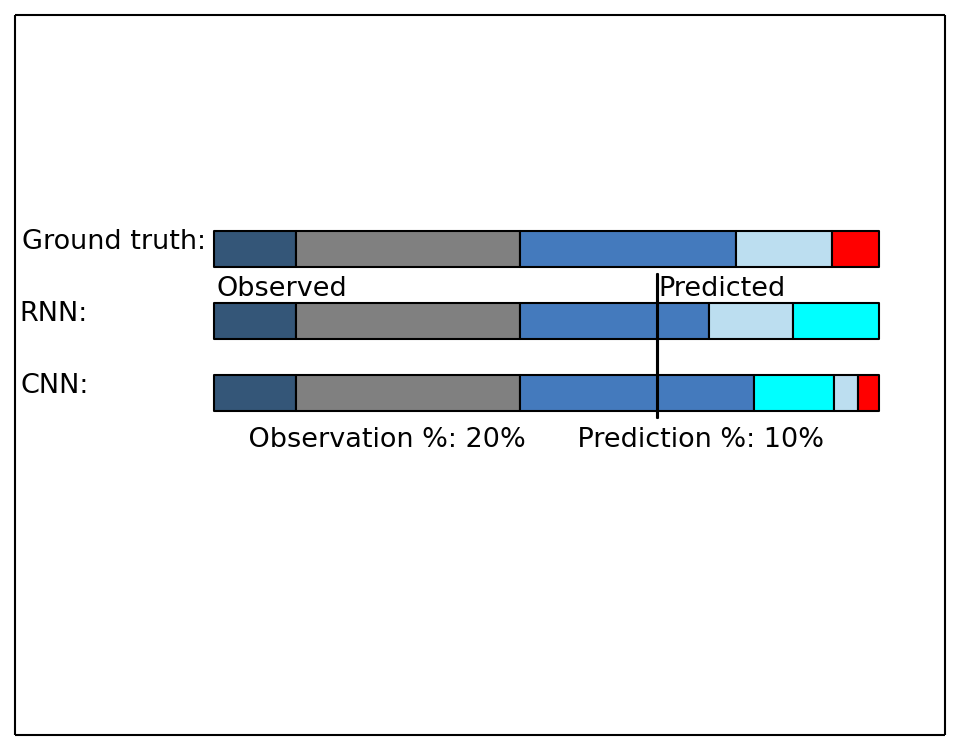}
& 
\includegraphics[trim={3mm 5cm 3mm 3cm},clip,width=.4\textwidth]{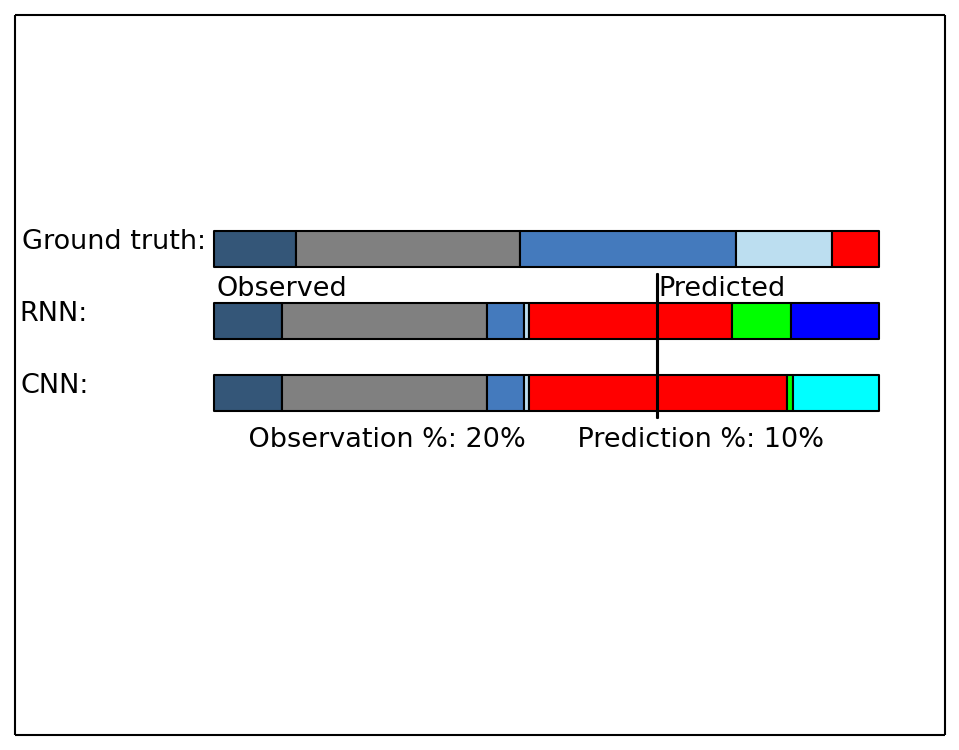}
\\
\begin{footnotesize}(c) Results on 50Salads with ground-truth observation \end{footnotesize}
& 
\begin{footnotesize}(d) Results on 50Salads without ground-truth observation \end{footnotesize}
\end{tabular}
\caption{Qualitative results for the future action prediction task for both, RNN and CNN with and without ground-truth observations. }
\label{fig:qual}
\vspace{-2mm} 
\end{figure*}

In Table~\ref{tab:gt_res} and Figure~\ref{fig:plots_gt}, the results on Breakfast and 50Salads
are shown.
Both the RNN model and the CNN model show good performance compared to the baseline.
Independent of the fraction of the video that has been observed, 
\ie $ 20\% $, or $ 30\% $, 
however, the RNN outperforms the CNN in most cases. In general, the RNN is better for short term prediction, 
\ie $10\%$, or $20\%$, while the CNN performs similarly or even sometimes better than the RNN for longer prediction. 
The reason lies in the recursive structure of the RNN predictions:
once a segment is predicted, it is appended to the observed part and used as input to predict
the next sequence. Consequently, if the RNN outputs an erroneous prediction at some point,
this error is likely to propagate through time. The CNN, on the contrary, uses the observed
part of the video to predict all future actions directly, so errors are less likely to
propagate from one segment to another. However, this slightly better performance of the CNN 
comes with the drawback of favouring long action segments over short ones.
The behaviour can also be observed in the qualitative results in Figure~\ref{fig:qual} (a) and (c).
For example in the case of (c), the CNN tends to miss small action segments, whereas the RNN seems to be more
reliable.




\begin{table}[t]
\caption{Results for future action prediction with ground-truth observations. Numbers represent accuracy as mean over classes.}
\label{tab:gt_res}
\resizebox{\columnwidth}{!}{%
\begin{tabular}{c|cccc|cccc}
\hline
Observation \% & 
\multicolumn{4}{c|}{\textbf{20\%}} &
\multicolumn{4}{c}{\textbf{30\%}} \\ \hline
Prediction \%  & 
\textbf{10\%} & \textbf{20\%} & \textbf{30\%} & \textbf{50\%} & 
\textbf{10\%} & \textbf{20\%} & \textbf{30\%} & \textbf{50\%}  \\ \hline

\multicolumn{9}{l}{\textbf{\textit{Breakfast}}} \\ \hline
Grammar &
0.4892		&		0.4033		&		0.3624		&		0.3146		&
0.5266		&		0.4215		&		0.3844		&		0.3309	
\\ 

Nearest-Neighbor      & 
0.4378      &       0.3726       &       0.3492      &       0.2984       &
0.4412      &       0.3769       &       0.3570      &       0.3019       
\\ 

RNN model      & 
\textbf{0.6035}  &  \textbf{0.5044}  &  \textbf{0.4528}  &  \textbf{0.4042}  &
\textbf{0.6145}  &  \textbf{0.5025}  &  0.4490           &  \textbf{0.4175}       
\\ 

CNN model      &
0.5797			&	0.4912			&	0.4403			&	0.3926			&
0.6032			&	0.5014			&	\textbf{0.4518}	&	0.4051	
\\ \hline

\multicolumn{9}{l}{\textbf{\textit{50Salads}}} \\ \hline
Grammar    &
0.2869			&	0.2165			&	0.1832			&	0.1037			&
0.2671			&	0.1459			&	0.1169			&	0.0925		
\\ 

Nearest-Neighbor      & 
0.2521          &   0.2105           &   0.1634           &  0.1317           &
0.2212          &   0.1715           &   0.1838           &  \textbf{0.1471}       
\\ 

RNN model      &
\textbf{0.4230}  &  \textbf{0.3119}  &  \textbf{0.2522}  &  \textbf{0.1682}  &
\textbf{0.4419}  &  \textbf{0.2951}  &  0.1996			&  0.1038	
\\ 

CNN model      & 
0.3608			&	0.2762			&	0.2143			&	0.1548			&
0.3736			&	0.2478			&	\textbf{0.2078}	&	0.1405
\\ \hline

\end{tabular}%
}
\end{table}

\subsection{Prediction without Ground-Truth Observations}

In this section, we evaluate the performance of our proposed systems given noisy annotations. In contrast to the clean ground-truth observations from the previous section, we now assume that the observed part of the video has been decoded using the system of~\cite{rnn_model} and, thus, $ \mathbf{c}_1^t $ is not perfect anymore but is likely to contain errors.
The mean over frames accuracy of~\cite{rnn_model} when observing $ 20\% $ of the video, for instance, is only $ 37\% $ on Breakfast 
and $ 67\% $ on 50Salads. While the accuracy when observing $ 30\% $ of the video is $ 43\% $ on Breakfast and $ 68\% $ on 50Salads.


\begin{table}[b]
\caption{Results for future action prediction without ground-truth observations. Numbers represent accuracy as mean over classes.}
\label{tab:noisy_res}
\resizebox{\columnwidth}{!}{%
\begin{tabular}{c|cccc|cccc}
\hline
Observation \% & 
\multicolumn{4}{c|}{\textbf{20\%}} & 
\multicolumn{4}{c}{\textbf{30\%}}  \\ \hline
Prediction \%  & 
\textbf{10\%} & \textbf{20\%} & \textbf{30\%} & \textbf{50\%} & 
\textbf{10\%} & \textbf{20\%} & \textbf{30\%} & \textbf{50\%} \\ \hline

\multicolumn{9}{l}{\textbf{\textit{Breakfast}}} \\ \hline
Grammar      & 
0.1660			&	0.1495			&	0.1347			&	0.1342			&
0.2110			&	0.1818			&	0.1746			&	0.1630				
\\ 

Nearest-Neighbor      & 
0.1642          &   0.1501           &   0.1447          &  0.1329   &
0.1988          &   0.1864           &   0.1797          &  0.1657       
\\ 

RNN model      & 
\textbf{0.1811}  &  \textbf{0.1720}  &  \textbf{0.1594}  &  \textbf{0.1581}  &
0.2164           &  0.2002           &  \textbf{0.1973}  &  \textbf{0.1921}  
\\ 

CNN model      & 
0.1790			&	0.1635			&	0.1537			&	0.1454			&
\textbf{0.2244}	&	\textbf{0.2012}	&	0.1969			&	0.1876			
\\ \hline

\multicolumn{9}{l}{\textbf{\textit{50Salads}}} \\ \hline
Grammar      & 
0.2473			&	0.2234			&	\textbf{0.1976}	&	0.1274			&
0.2965			&	0.1918			&	0.1517			&	0.1314
\\ 

Nearest-Neighbor      & 
0.1904          &   0.1610           &   0.1413          &  0.1037  &
0.2163          &   0.1548           &   0.1347          &  \textbf{0.1390}       
\\ 

RNN model      & 
\textbf{0.3006}	&	\textbf{0.2543}	&	0.1874       	&	\textbf{0.1349}	&
\textbf{0.3077}	&	0.1719			&	0.1479			&	0.0977			
\\ 

CNN model      & 
0.2124			&	0.1903			&	0.1598			&	0.0987			&
0.2914			&	\textbf{0.2014}	&	\textbf{0.1746}	&	0.1086		
\\ \hline

\end{tabular}%
}
\end{table}

\begin{figure}[t]
\centering
\begin{tabular}{cc}
\includegraphics[width=.45\columnwidth]{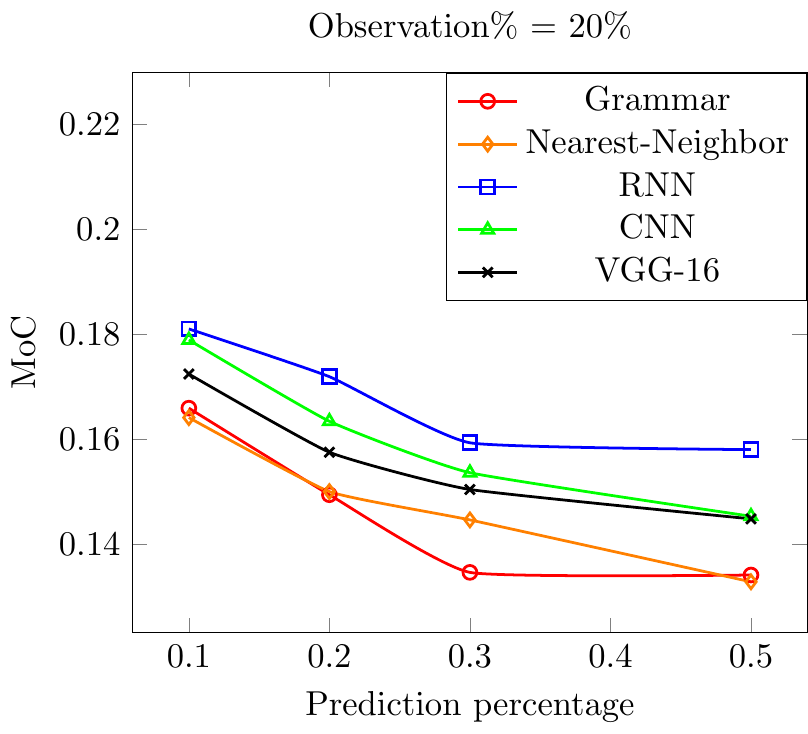}
&
\includegraphics[width=.45\columnwidth]{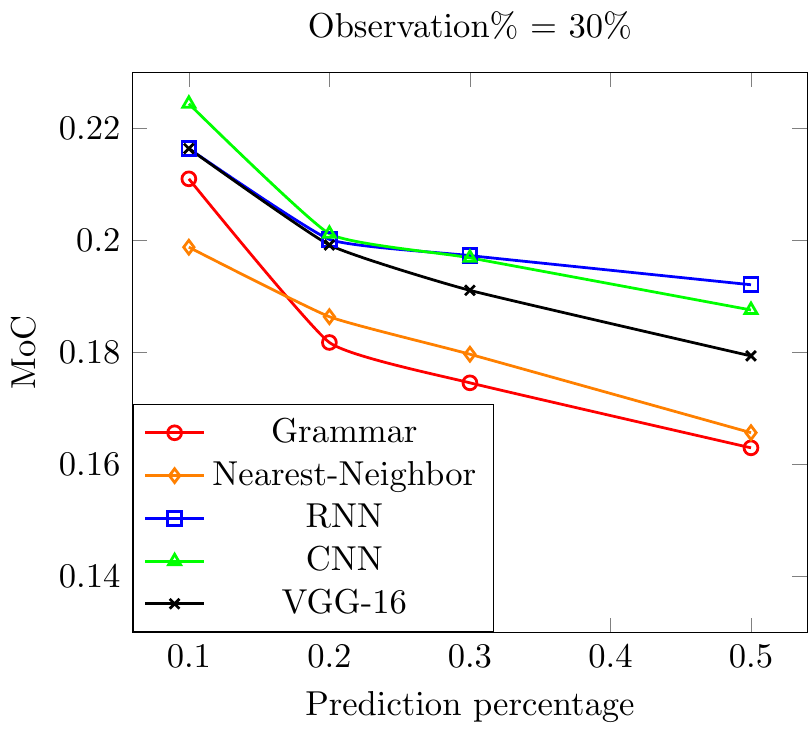} 
\\
\multicolumn{2}{c}{
\begin{footnotesize}
(a) Results on Breakfast
\end{footnotesize}
}
\\
\includegraphics[width=.45\columnwidth]{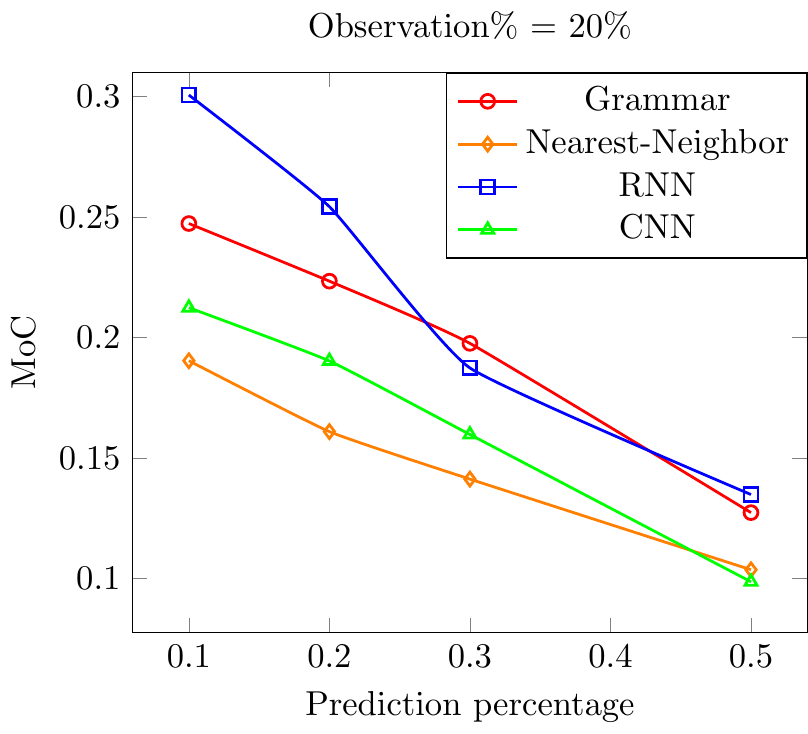}
&
\includegraphics[width=.45\columnwidth]{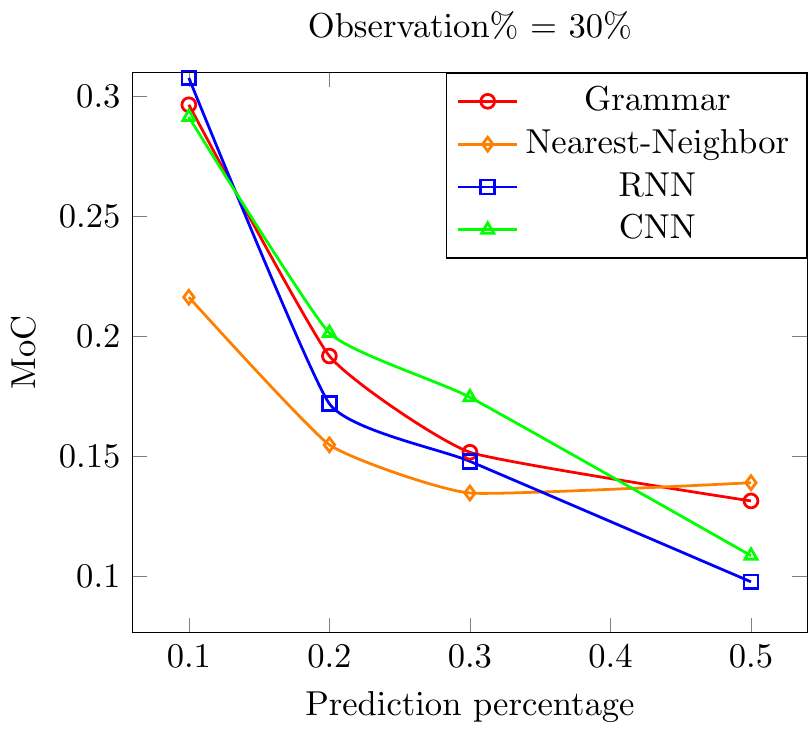} 
\\
\multicolumn{2}{c}{
\begin{footnotesize}
(b) Results on 50Salads
\end{footnotesize}
}
\end{tabular}
\caption{Results for future action prediction without ground-truth observations. }
\label{fig:plots_noisy}
\vspace{-2mm} 
\end{figure}

Compared to the amount of noise in the observed video labeling, the prediction results are still surprisingly stable, see Table~\ref{tab:noisy_res} and Figure~\ref{fig:plots_noisy}.
While the drop in performance on the Breakfast dataset (Table~\ref{tab:gt_res} vs.\ Table~\ref{tab:noisy_res}) is comparably large, both systems still
achieve a good performance compared to the 
baseline. 
On 50Salads, the overall loss of accuracy compared to the system with perfect observations is surprisingly small.
This can on the one hand be attributed to the better performance of the decoder~\cite{rnn_model}. On the other hand, the inter-class dependencies
on 50Salads are very strong, making it easier for both RNN and CNN to learn valid action sequences.

We would also like to put emphasis on the qualitative results for noisy observations, see Figure~\ref{fig:qual} (b) and (d).
Particularly, the case of (d) is interesting: the decoder mistakenly predicted incorrect label for the last observed segment.
For both models, RNN and CNN, this error propagates further to future segment predictions.

\begin{figure}[t]
    \centering
    \includegraphics[width=.4\textwidth]{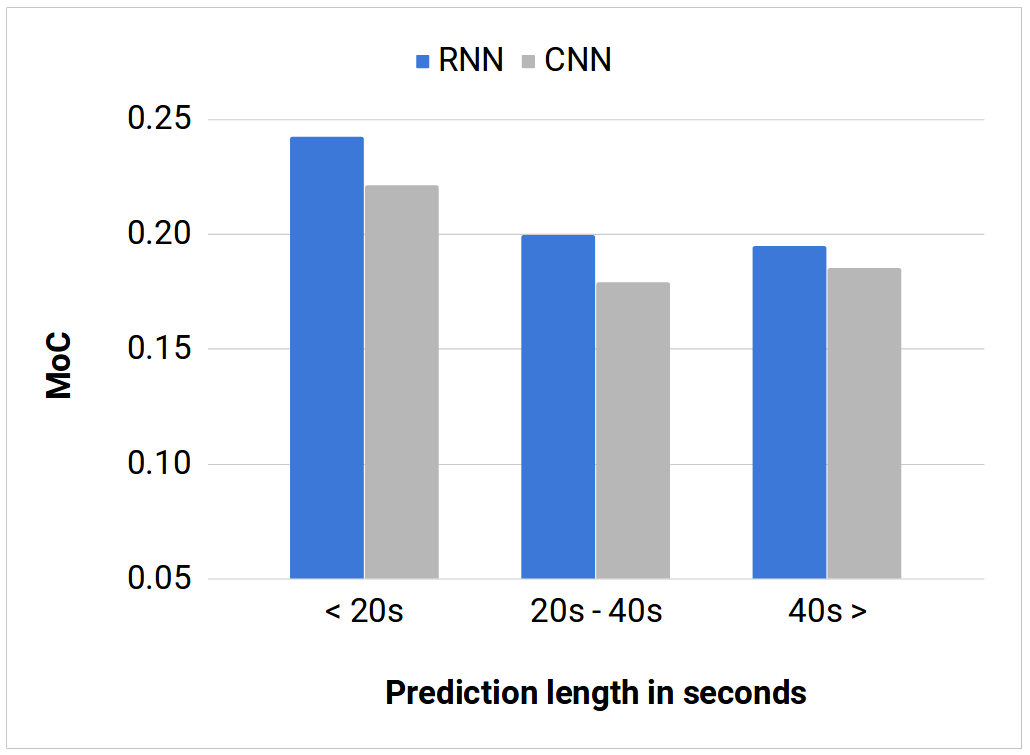}
    \caption{Performance of the models on videos with different lengths from the Breakfast dataset without ground-truth observations for the case of observing $30\%$ of the videos and predicting the following $50\%$.}
    \label{fig:vid_len}
\end{figure}

As the videos have different lengths, the proposed models might behave differently depending on the length of the videos.
An evaluation on three categories of videos from Breakfast based on the prediction length is shown in Figure~\ref{fig:vid_len} for the case of observing $ 30\% $ of the video and predicting the $ 50\% $ that follows. 
While the models perform well on both short and long videos, they achieve a better performance for shorter videos. 
This is mainly because the duration of the predicted future is less for shorter videos, which makes the 
prediction task much easier compared to long video sequences.
A similar behaviour can also be observed when considering the number of actions that are predicted in the future.
As shown in Table~\ref{tab:actions_res}, the accuracy drops as we predict more in the future.

\begin{table}[t]
\centering
\caption{Accuracy over next, 2nd action, and 3rd action for 30\% observation and 50\% prediction for Breakfast without ground-truth observations. The action is correctly detected if the IoU of the predicted action segment with the annotated action segment is $\geqslant 0.5$.}
\label{tab:actions_res}
\resizebox{0.7\columnwidth}{!}{%
\begin{tabular}{cccc}
\hline
& 1st action & 2nd action & 3rd action\\
\hline
Grammar &  0.2232  &  0.1325  &  0.1117  \\
Nearest-Neighbor &  0.2295  &  0.1323  &  0.1234  \\
RNN      &  \textbf{0.2643}  & \textbf{ 0.1773 } &  \textbf{0.1792}   \\
CNN      &  0.2595  &  0.1603  &  0.1425  
\\ \hline
\vspace{-4mm} 
\end{tabular}%
}
\end{table}

\subsection{Analysis of the CNN Model}
\paragraph{Model Architecture}
Compared to commonly used CNN architectures such as VGG-16 or ResNet, our model is comparably small.
We stick to this simple architecture since we have very limited amount of training data, and complex models would easily overfit on the training set.
A comparison of our architecture to a VGG-16 \cite{vgg} on Breakfast with and without ground truth annotation is provided in Figure~\ref{fig:plots_gt} and Figure~\ref{fig:plots_noisy}, respectively.
The performance of our architecture compared to VGG-16 is clearly better with an improvement up to $9\%$ 
when using the ground truth annotation as observations.
Note that our architecture already has around $ 6 $m parameters due to the fully connected layers at the end.

\begin{figure*}[t]
\centering
\begin{tabular}{cc}
\includegraphics[trim={3mm 5cm 3mm 4.6cm},clip,width=.4\textwidth]{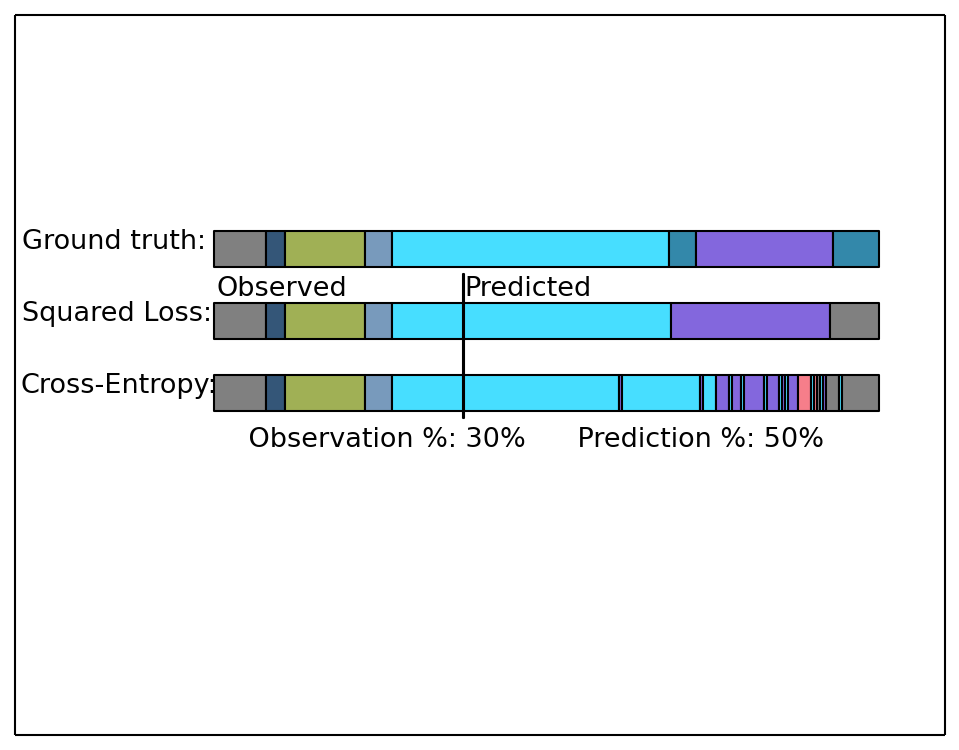} 
& 
\includegraphics[trim={3mm 5cm 3mm 4.6cm},clip,width=.4\textwidth]{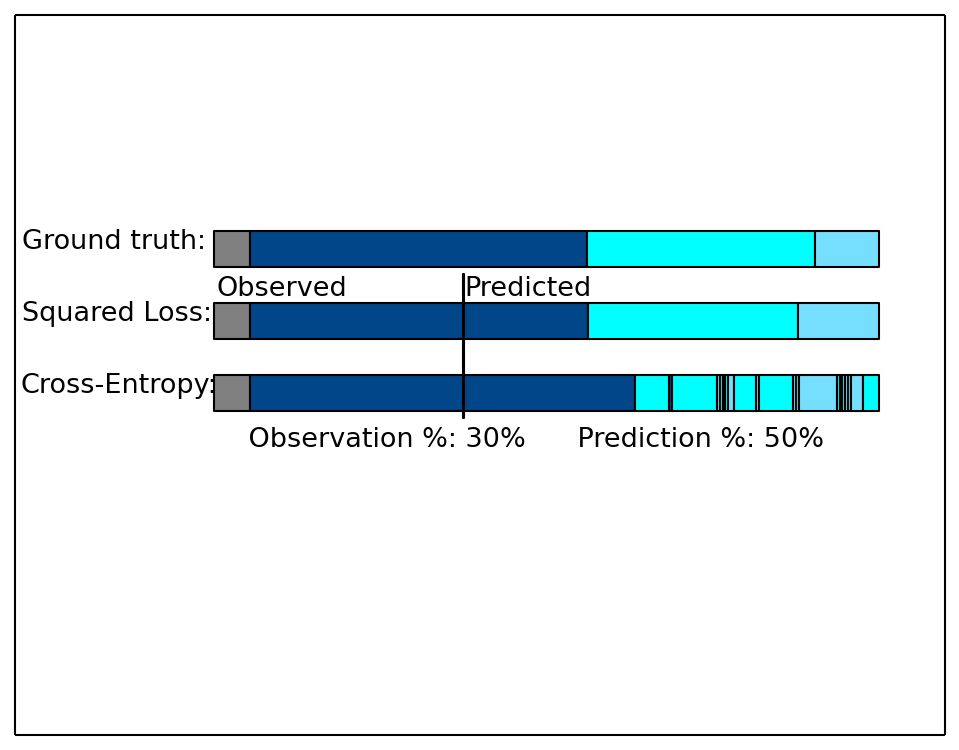}
\\
\end{tabular}
\caption{Results of the CNN model using the cross-entropy loss vs.\ the squared-error loss.}
\label{fig:softmaxVsL2}
\vspace{-2mm} 
\end{figure*}

\begin{figure*}[t]
\centering
\begin{tabular}{cc}
\includegraphics[trim={3mm 5cm 3mm 4.6cm},clip,width=.4\textwidth]{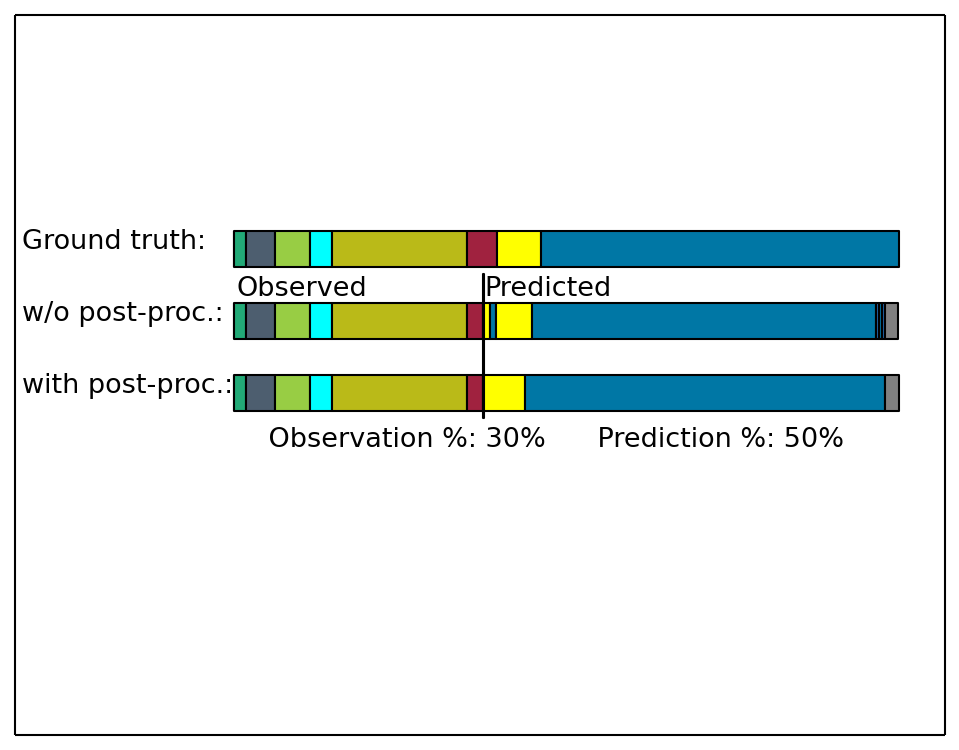} 
& 
\includegraphics[trim={3mm 5cm 3mm 4.6cm},clip,width=.4\textwidth]{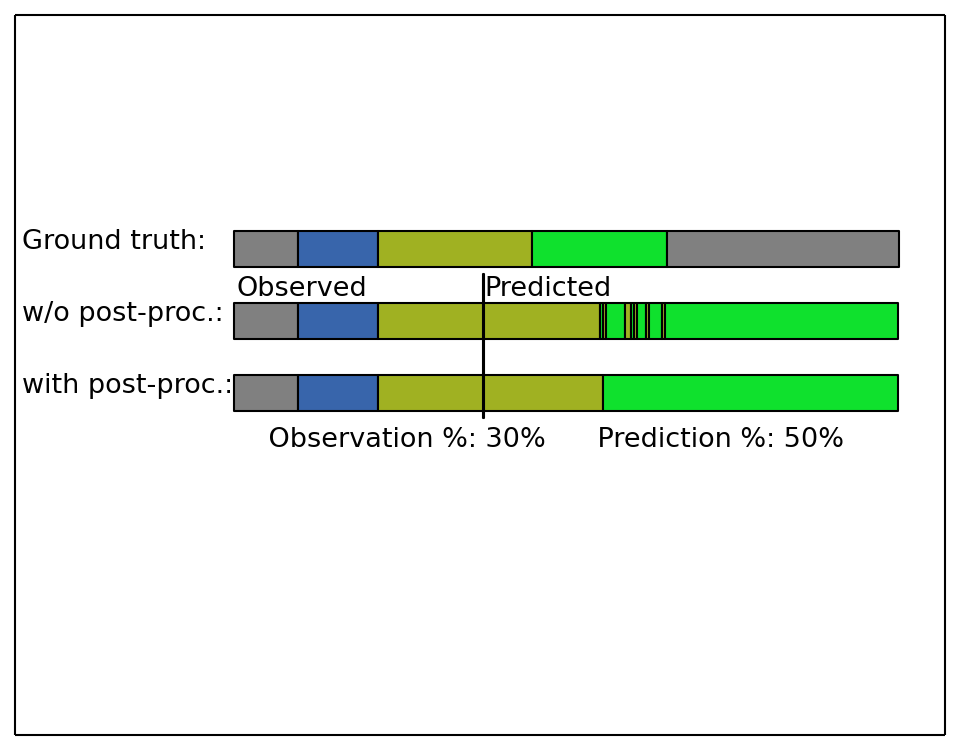}
\end{tabular}
\caption{Results of the CNN model with and without post-processing.}
\label{fig:post_proc}
\vspace{-2mm} 
\end{figure*}

\paragraph{Loss Function}
Our choice of the loss function for the CNN based model is a squared loss preceded by an $ \ell_2 $ normalization layer. However, for classification tasks, a softmax with cross-entropy loss is usually used. Since the future action prediction task can be viewed as special kind of classification, a softmax layer trained with the cross-entropy loss seems a more suitable choice on first glance.
Yet, a comparison of both loss types shows a clear superiority of the squared loss combined with the $ \ell_2 $ normalization layer, see Table~\ref{tab:cnn_analysis}. We attribute this to the smoothing properties of the softmax layer. Even large differences of the input values to a softmax layer
lead to comparably small differences in the output probability distribution. This is a desirable effect in classification tasks.
For our task, however, it leads to frequent changes of the maximizing label index, which corresponds to over-segmentations.
The effect can also be observed in Figure~\ref{fig:softmaxVsL2}. While strong over-segmentation can be observed for the
softmax output, this effect nearly completely vanishes using the $ \ell_2 $ normalization layer. Nevertheless, even with the $ \ell_2 $ normalization layer, a small over-segmentation effect is still visible in some cases, which can be eliminated by the post-processing step as shown in Figure~\ref{fig:post_proc}.

\begin{table}[t]
\caption{Comparing different loss functions for the CNN model on Breakfast with ground-truth observations. Numbers represent accuracy as mean over classes.}
\label{tab:cnn_analysis}
\resizebox{\columnwidth}{!}{%
\begin{tabular}{c|cccc|cccc}
\hline
Observation \%   & 
\multicolumn{4}{c|}{\textbf{20\%}}      & 
\multicolumn{4}{c}{\textbf{30\%}}    \\ \hline
Prediction \%    & 
\textbf{10\%} & \textbf{20\%} & \textbf{30\%} & \textbf{50\%} & 
\textbf{10\%} & \textbf{20\%} & \textbf{30\%} & \textbf{50\%}  \\ 
\hline

CNN (cross-entropy)   &
0.5177			&	0.4280			&	0.3940			&	0.3625			&
0.5499			&	0.4838			&	0.4324			&	0.3858		
\\ 

CNN (squared loss)      & 
\textbf{0.5797}	&	\textbf{0.4912}	&	\textbf{0.4403}	&	\textbf{0.3926}	&
\textbf{0.6032}	&	\textbf{0.5014}	&	\textbf{0.4518}	&	\textbf{0.4051}
\\ \hline

\end{tabular}%
}
\end{table}

\subsection{Future Prediction Directly from Features}

So far, we considered observations that are either frame-wise ground truth action labels, or those labels that are obtained by decoding the observed frame-wise features. In this section, we evaluate the performance of our models when applied directly on the observed frame-wise features and compare it to the two-step approach. 
We only use the CNN model for this evaluation. Originally, the input of the model is a matrix with $C$ columns that correspond to the number of classes, and $S$ rows that represent the temporal resolution. Each row is a 1-hot encoding that represents the action label of the corresponding frames in the observed sequence. When applying this model to features directly, $C$ is equal to the dimensionality of the features, which is 64 in our case for the Fisher vectors features. 
$ S $ is kept at $ 128 $ as before. The observed video features are down- or upsampled to have exactly 128 frames. We use the same training protocol that is used for the previous experiments, by considering the set $\{10\%,\ 20\%,\ 30\%,\ 50\%\}$ as observation percentages, and the target is always the $50\%$ that follows immediately after the observations. Table~\ref{tab:features} shows the results of the CNN model when applied on features directly compared to the two-step approach. As shown in the table, using the two-step approach outperforms the direct prediction from features by a large margin, \ie up to $+5\%$. Predicting the future directly from features is a harder problem since the model has to recognize the observed actions and capture the relevant information to anticipate the future, while for the two-step approach these two tasks are decoupled. This allows to use a strong decoding model to recognize the observed actions, and restricting the future predictor to capture the context over action classes only instead of frame-wise features. A similar conclusion was reached by \cite{martin} where semantic labels of the unseen parts are predicted in the spatial domain of an image. It has been shown that segmenting the observed part of the image first and then using the segmented image for prediction achieves better results than using the RGB image directly.

\begin{table}[!htp]
\caption{Results for future action prediction directly from features on the Breakfast dataset. Numbers represent accuracy as mean over classes.}
\label{tab:features}
\resizebox{\columnwidth}{!}{%
\begin{tabular}{c|cccc|cccc}
\hline
Observation \%   & 
\multicolumn{4}{c|}{\textbf{20\%}}         & 
\multicolumn{4}{c}{\textbf{30\%}}     \\ \hline
Prediction \%    & 
\textbf{10\%} & \textbf{20\%} & \textbf{30\%} & \textbf{50\%} & 
\textbf{10\%} & \textbf{20\%} & \textbf{30\%} & \textbf{50\%} \\ \hline

CNN (features)   & 
0.1278			&	0.1162			&	0.1121			&	0.1027			&
0.1772			&	0.1687			&	0.1548			&	0.1409		
\\ 

CNN (w/o GT obs.) & 
\textbf{0.1790}	&	\textbf{0.1635}	&	\textbf{0.1537}	&	\textbf{0.1454}	&
\textbf{0.2244}	&	\textbf{0.2012}	&	\textbf{0.1969}	&	\textbf{0.1876}	
\\ \hline
\end{tabular}%
}
\end{table}

%
%

\subsection{Comparison with the State-of-the-Art}

To the best of our knowledge, long term future action prediction has not been addressed before. Most works focus on predicting the immediate future.
Vondrick \etal~\cite{vondrick}, for instance, train a model to predict AlexNet features of a frame one second in the future.
Based on these predictions, an SVM is trained to determine the action label of the future frame.

Since~\cite{vondrick} train their future prediction model on $ 600h $ of videos from the web, which are not made publicly available,
we use the recent Kinetics network to generate deep CNN features that have shown to generalize extremely well on several action
recognition datasets and are the current state-of-the-art~\cite{kinetics}.
We run the approach of~\cite{vondrick} on both, Breakfast and 50Salads, and compare their results to our model.
To provide a fair comparison, our model is trained in a way that the input sequences always
end one second before the next action segment starts. The results are shown in Table~\ref{tab:state_of_the_art}.
Our approach outperforms the system of~\cite{vondrick} by a large margin. Note that for both datasets,
predicting an action label only based on a single frame is particularly hard and the framewise action classifier
achieves less than $ 10\% $ accuracy.
In contrast to~\cite{vondrick}, our approaches make use of temporal context of the previously observed video content, which is crucial for reliable predictions.

\begin{table}[ht]
\small
\centering
\caption{Comparison with~\cite{vondrick}: Accuracy of predicting future actions one second before they start is reported.}
\label{tab:state_of_the_art}
\resizebox{0.7\columnwidth}{!}{%
\begin{tabular}{lcc}
\hline
                 & Breakfast          & 50Salads \\ \hline 
Vondrick \etal~\cite{vondrick}  & 8.1                & 6.2      \\ 
RNN model        & \textbf{30.1}      & \textbf{30.1}      \\ 
CNN model        & 27                 & 29.8       \\ \hline
\end{tabular}
}%
\end{table}

\section{Conclusion}
\label{sec:conclusion}

We have introduced two efficient methods to predict future
actions in videos, which is a task that has not been addressed before.
While most existing prediction approaches focus on early anticipation of an already
ongoing action or predict at most one action in the future, our methods are the first
to predict video content of up to several minutes length.
Proposing two models to address the task, an RNN and a CNN, we obtain
accurate predictions that scale well along different datasets and videos with
varying lengths, varying quality of observed data, and huge variations in the
possible future actions.

\paragraph{Acknowledgements:} The work has been financially supported by the DFG project GA 1927/4-1 
(DFG Research Unit FOR 2535 Anticipating Human Behavior), 
the ERC Starting Grant ARCA (677650) and the German  Academic Exchange Service (DAAD).
{\small
\bibliographystyle{ieee}
\bibliography{references}
}

\end{document}